\documentclass[runningheads]{llncs}
\usepackage{graphicx}

\usepackage{tikz}
\usepackage{comment}
\usepackage{amsmath,amssymb} 
\usepackage{color}
\usepackage{makecell}
\usepackage{subfig}
\usepackage{hyperref} 
\usepackage{booktabs}

\usepackage[accsupp]{axessibility}  

\begin{document}
\pagestyle{headings}
\mainmatter
\def\ECCVSubNumber{7274}  

\title{CardioSyntax: end-to-end SYNTAX score prediction \textendash~dataset, benchmark and method} 



\author{Alexander Ponomarchuk\inst{1} \and
Ivan Kruzhilov\inst{1} \and
Galina Zubkova\inst{1} \and Artem Shadrin\inst{2}  \and Ruslan Utegenov\inst{2}  \and Ivan Bessonov\inst{2}  \and Pavel Blinov\inst{1}}

\authorrunning{A Ponomarchuk et al.}

\institute{Sber AI Lab  \and
Tyumen Cardiology Research Center, Tomsk National Research Medical Center of Russian Academy of Science\\
}
\maketitle

\begin{abstract}
  The SYNTAX score has become a widely used measure of coronary disease severity, crucial in selecting the optimal mode of the revascularization procedure. This paper introduces a new medical regression and classification problem — automatically estimating SYNTAX score from coronary angiography. Our study presents a comprehensive CardioSYNTAX dataset\footnote{The dataset is avaliable at \href{https://zenodo.org/records/14005818}{\url{https://zenodo.org/records/14005818}}} of 3,018 patients for the SYNTAX score estimation and coronary dominance classification. The dataset features a balanced distribution of individuals with zero and non-zero scores. This dataset includes a first-of-its-kind, complete coronary angiography samples captured through a multi-view X-ray video, allowing one to observe coronary arteries from multiple perspectives. Furthermore, we present a novel, fully automatic end-to-end method for estimating the SYNTAX. For such a difficult task, we have achieved a solid coefficient of determination $R^{2}$ of 0.51 in score value prediction and 77.3\% accuracy for zero score classification.
  
\end{abstract}

\section{Introduction\protect\footnote{The article is also available at \href{doi: 10.1109/WACV61041.2025.00573}{\url{https://doi.org/10.1109/WACV61041.2025.00573}}. Please cite the paper as: \textit{A. Ponomarchuk et al., "CardioSyntax: End-to-End SYNTAX Score Prediction - Dataset, Benchmark and Method," 2025 IEEE/CVF Winter Conference on Applications of Computer Vision (WACV), Tucson, AZ, USA, 2025, pp. 5873-5883.}}}

\begin{figure*}[]
\centering
\includegraphics[height=5.4cm]{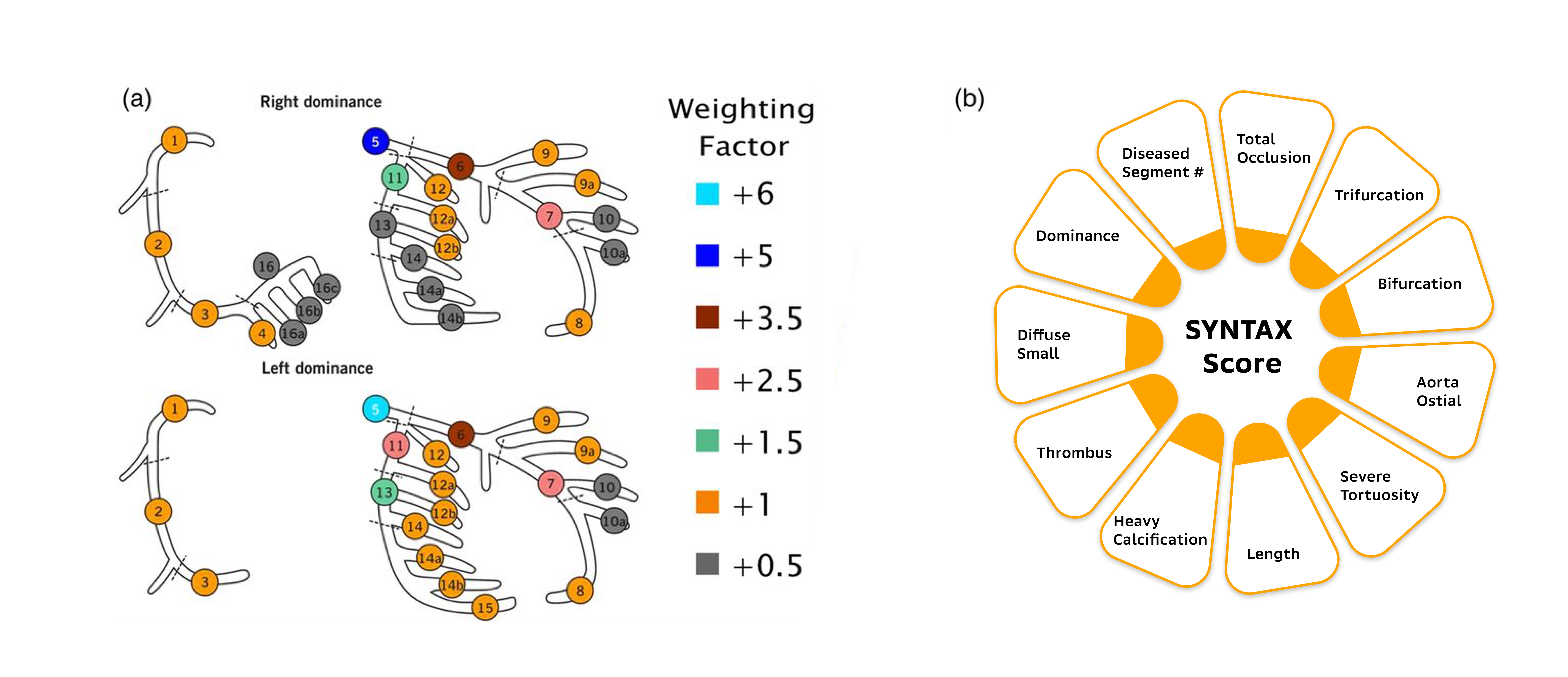}
\caption{SYNTAX score evaluation scheme. (a) Weighting factors for artery branches~\cite{farooq2013widening}. (b) Constituent factors of SYNTAX score.}
\label{fig:syntax_scheme}
\end{figure*}

Nowadays, heart disease is the leading cause of death in all developed societies~\cite{di2023world}. Coronary angiography is the primary visualization method used to examine a diseased heart. The cardiology community proposed a particular score to quantify and comprehensively assess the severity of the damage to the heart's coronary vessels from angiography. The SYNTAX (SYNergy between percutaneous coronary intervention with TAXus and cardiac surgery) score is a widely used~\cite{lawton20222021} and validated risk assessment tool for selecting revascularization strategies in patients with multi-vessel coronary disease~\cite{sianos2005syntax}. Using the SYNTAX score has proven beneficial in clinical decision-making and has been widely adopted in medical research and clinical practice~\cite{akboga2023systemic,gaudino2023comparison}. The use of the SYNTAX scale has been included in the recommendations from the European Society of Cardiology~\cite{vrints20242024}.

However, despite the widespread use of the SYNTAX score, there is still no fully automatic tool for its calculation. Cardiologists continue using semi-automatic SYNTAX calculator tools in their daily work, which are highly subjective, time-consuming, and error-prone. As noted by~\cite{basman2022variability}, there is substantial variability in the scoring among different physicians due to the meticulous work involved. A fully automatic SYNTAX score prediction could address these issues and substantially boost the cardiologists' performance.

\begin{figure*}
\centering
\subfloat[]{\includegraphics[height=3.0cm]{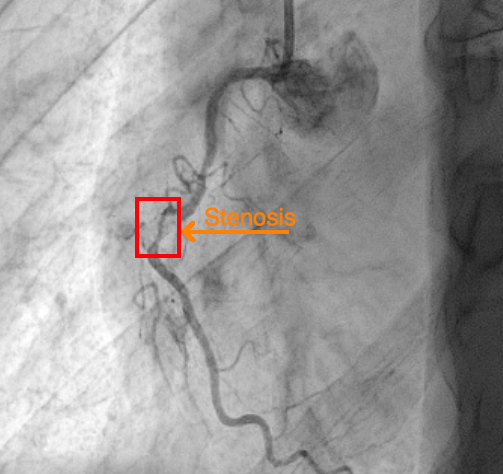} \label{fig:fig_stenosis}}
\subfloat[]{\includegraphics[height=3.0cm]{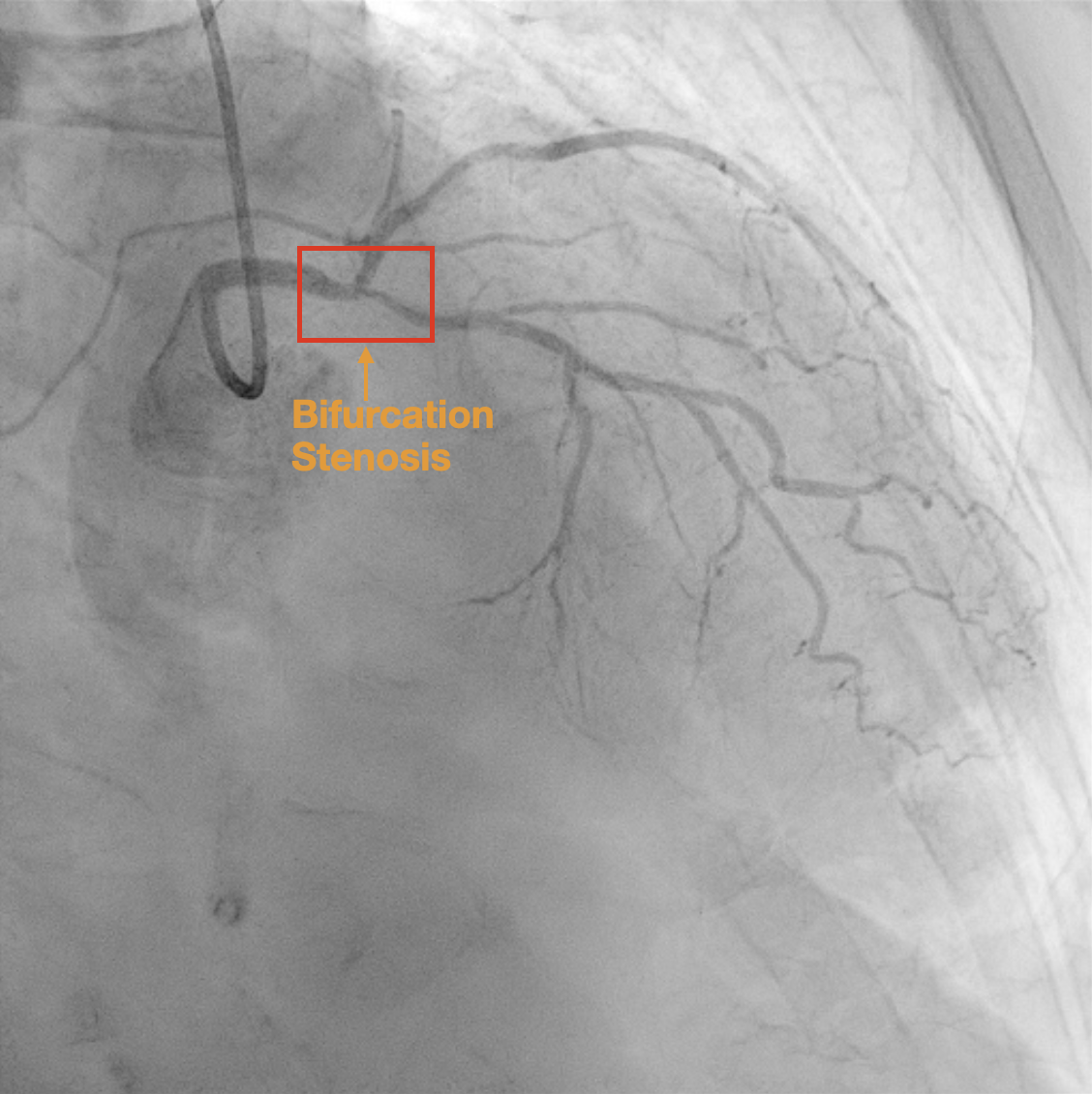} \label{fig:fig_bifurcation}}
\subfloat[]{\includegraphics[height=3.0cm]{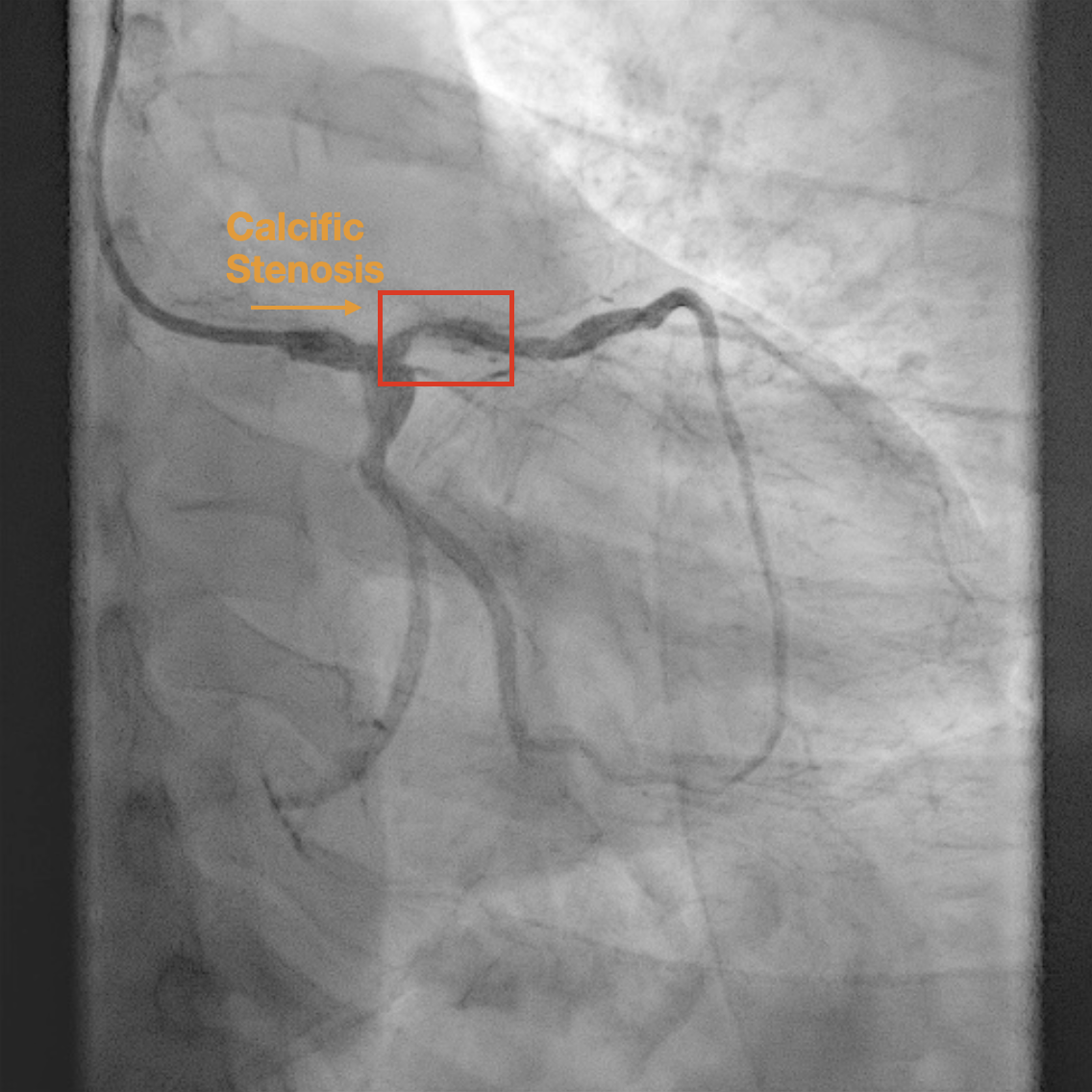} \label{fig:fig_calcific}}
\subfloat[]{\includegraphics[height=3.0cm]{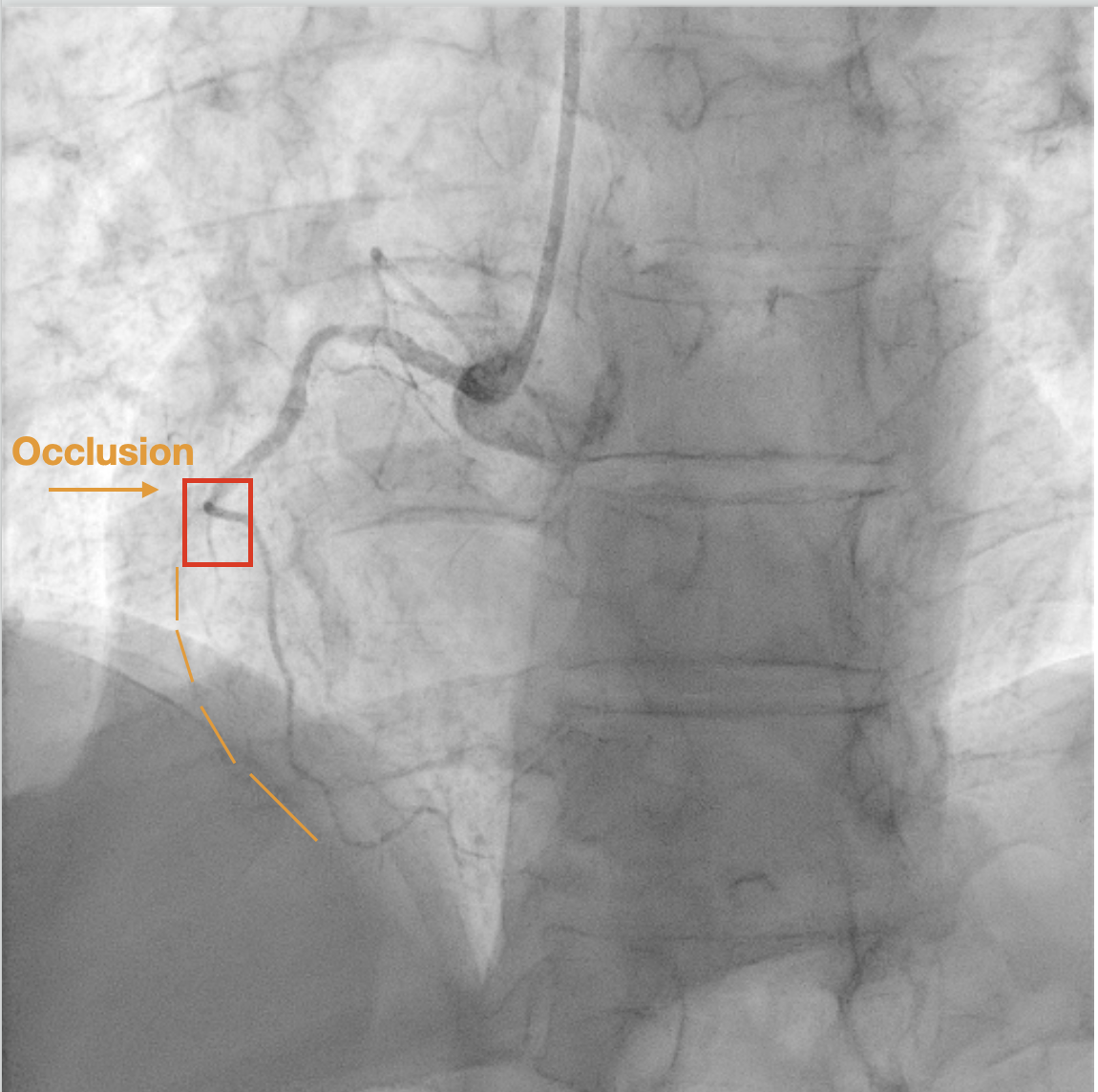} \label{fig:fig_occlusion}}
\caption{An example of anomalies in right coronary artery. (a) Stenosis. (b) Bifurcation. (c) Calcific stenosis. (d) Total occlusion.}
\label{fig:example_stenosis}
\end{figure*}

The SYNTAX score estimation is a multi-step process, as shown in Fig.~\ref{fig:syntax_scheme}. This scoring system considers various factors, such as the number and complexity of lesions, thrombus presence, etc. The score estimation begins with classifying coronary dominance type, whether right or left, as the downstream weighting factors depend on the dominance type.
The next step involves selecting the artery segment with stenosis (Fig.~\ref{fig:fig_stenosis}) or occlusion (Fig.~\ref{fig:fig_occlusion}), followed by indicating additional characteristics \textendash~calcification (Fig.~\ref{fig:fig_calcific}), bifurcation (Fig.~\ref{fig:fig_bifurcation}), etc.
Finally, the significance scores~\cite{sianos2005syntax} assigned to the lesions are multiplied by weighting factors, shown in Fig.~\ref{fig:syntax_scheme}a, and summed up to the total SYNTAX score. It is worth noting that a higher numeric value of an artery branch indicates a deeper position or lower importance in the coronary tree, with subsequent segments always having a higher number than the segments they originate from.

Correctly establishing coronary arterial dominance is essential as a base for the downstream steps. There are two main dominance categories: left and right ~\cite{shriki2012identifying}. Approximately 70-80\% of the population has right coronary dominance, while about 5-10\% have left coronary dominance, and the rest have a co-dominance. A specific coronary artery branch 
determines the dominance type. It is hard to detect such a detail in a coronary tree image, leading to costly errors. If this preliminary step is mistaken, the error in the SYNTAX score can exceed 30\%.

A raw angiographic study is a dynamic X-ray video of the heart obtained by injecting a contrast agent into a coronary artery. That means that typical angiography includes several views. Each view captures the heart's pulsation differently, allowing it to state the SYNTAX score estimation as a multi-view image processing problem. Since not all projections are equally informative, it is essential to intelligently aggregate information from them.

Together, these factors make the problem of automatic SYNTAX estimation challenging from a data point of view and non-trivial for current methods in computer vision.

Our main contributions can be summarized as follows:
\begin{itemize}
\item We introduce CardioSYNTAX, a new and extensive coronary angiogram dataset of 3,018 patients. This dataset features multi-view X-ray videos and their corresponding SYNTAX scores. This dataset is the first collection of X-ray coronary angiograms to date, providing a valuable resource for future research.

\item We establish a novel benchmark for SYNTAX score estimation and coronary dominance classification tasks. Two interventional cardiologists independently labeled 60 studies from our dataset, which allows us to estimate the intra-level disagreement in SYNTAX score prediction ($R^2\approx0.7$ and Bland-Altman plot $STD \approx8.5$).

\item We propose a novel, fully automatic end-to-end Multi-View Look (MVL) method for estimating SYNTAX scores using X-ray coronary angiography. Our method achieves a solid $R^2$ of 0.51 compared to experts, with great potential for further research and improvement and 77.3\% accuracy for zero score classification.
\end{itemize}

\section{Related works}
When following well-established guidelines~\cite{sianos2005syntax} for SYNTAX score estimation, the steps needed to perform are:
\begin{itemize}
\item RCA (Right Coronary Artery) and LCA classification (Left Coronary Artery),
\item  Coronary dominance classification,
\item  Segmentation of arteries and their branches,
\item  Stenosis or occlusion detection,
\item  Estimation of specific characteristics,
\item  Final calculation of the score.
\end{itemize}

The studies~\cite{eschen2022classification,moalla2023exploiting} have successfully addressed the problem of RCA and LCA classification, achieving an impressive F1-score of 99\%. Building on this progress, the work~\cite{kruzhilov2023neural} has further contributed to the field by developing a method for classifying coronary dominance using only the RCA with a Recall macro above 90\%. However, to improve the classification quality further, the authors have suggested incorporating additional information, such as labeling occlusions and utilizing LCA data. In particular, several studies have proposed various artery segmentation~\cite{cervantes2019automatic,iyer2021angionet,nobre2023coronary,park2023selective,tao2022lightweight,wang2020coronary,yang2019deep,zhao2020semantic} and stenosis detection techniques~\cite{avram2023cathai,cong2023deep,danilov2021real,kavipriya2023identification,rodrigues2021automated}.

\subsection{SYNTAX datasets}
Using publicly available datasets is crucial for advancing research in computer vision methods for coronary artery disease. However, our analysis of this topic has revealed that most studies rely on private datasets, with only three are publicly available and non of them contains SYNTAX. One study~\cite{cervantes2019automatic} published 134 X-ray coronary frames with segmented arteries, each represented by a frame with 300*300 resolution. Another study~\cite{danilov2021real} provided a dataset of 8,325 angiogram frames with labeled stenosis areas collected from 100 patients using Siemens and GE cardiovascular imaging systems, with a frame resolution 512*512. The most recent study~\cite{popov2024dataset} introduced an ARCAD dataset containing 1500 frames for stenosis detection and 1500 frames segmented into 26 classes according to the SYNTAX description, with a resolution 512*512. It is worth noting that this dataset includes no more than 12 frames from a single patient and is acquired from two different cardio vascular imaging systems - Philips Azurion 3 and Siemens Artis Zee.

The scarcity and selective nature of publicly available datasets and the lack of shared code with pre-trained weights make it challenging to integrate the previous study into a single pipeline for estimating the SYNTAX score.

\subsection{SYNTAX datasets}
Using publicly available datasets is crucial for advancing research in computer vision methods for coronary artery disease. However, our analysis of this topic has revealed that most studies rely on private datasets, with only three are publicly available and non of them contains SYNTAX. One study~\cite{cervantes2019automatic} published 134 X-ray coronary frames with segmented arteries, each represented by a frame with 300*300 resolution. Another study~\cite{danilov2021real} provided a dataset of 8,325 angiogram frames with labeled stenosis areas collected from 100 patients using Siemens and GE cardiovascular imaging systems, with a frame resolution 512*512. The study~\cite{popov2024dataset} introduced an ARCAD dataset containing 1,500 frames for stenosis detection and 1,500 frames segmented into 26 classes according to the SYNTAX description, with a resolution 512*512. It is worth noting that this dataset includes no more than 12 frames from a single patient and is acquired from two different cardio vascular imaging systems \textendash~Philips Azurion 3 and Siemens Artis Zee. The CADICA dataset~\cite{jimenez2024cadica} contains full invasive coronary angiography and related meta-data for only 42 patients.

The scarcity and selective nature of publicly available datasets and the lack of shared code with pre-trained weights make it challenging to integrate the previous study into a single pipeline for estimating the SYNTAX score. The public CardioSYNTAX dataset of 3,018 full angiographic studies proposed in this work solves the aforementioned problems.

End-to-end estimation is an alternative to the step-by-step approach to estimating SYNTAX scores. To our knowledge, there is not researches attempted to estimate a SYNTAX score using angiograms. However, two approaches try to estimate SYNTAX scores based on medical tables. The study~\cite{ainiwaer2023machine} used XGBoost to predict SYNTAX scores based on 53 variables: age, sex, smoking status, brain natriuretic peptide level, left ventricular ejection fraction, etc. The study found that the model achieved an accuracy of approximately 7.0 standard deviations and an $R^2$ value of 0.54 compared to the ground truth, in their study~\cite{panteris2022machine}, used demographic, clinical, biochemical, protein, and metabolite variables to estimate the severity of coronary artery disease (CAD). They classified patients into two categories based on their SYNTAX scores: those with scores greater than one and those with less than or equal to 1. The ROC AUC for this classification reached 72.5\%, indicating satisfactory performance distinguishing between the two groups. Although such noninvasive methods help predict CAD risk, they cannot be applicable to determine the best mode of revascularization.

\section{CardioSYNTAX dataset}

\begin{figure}
\centering
\includegraphics[width=3.4in]{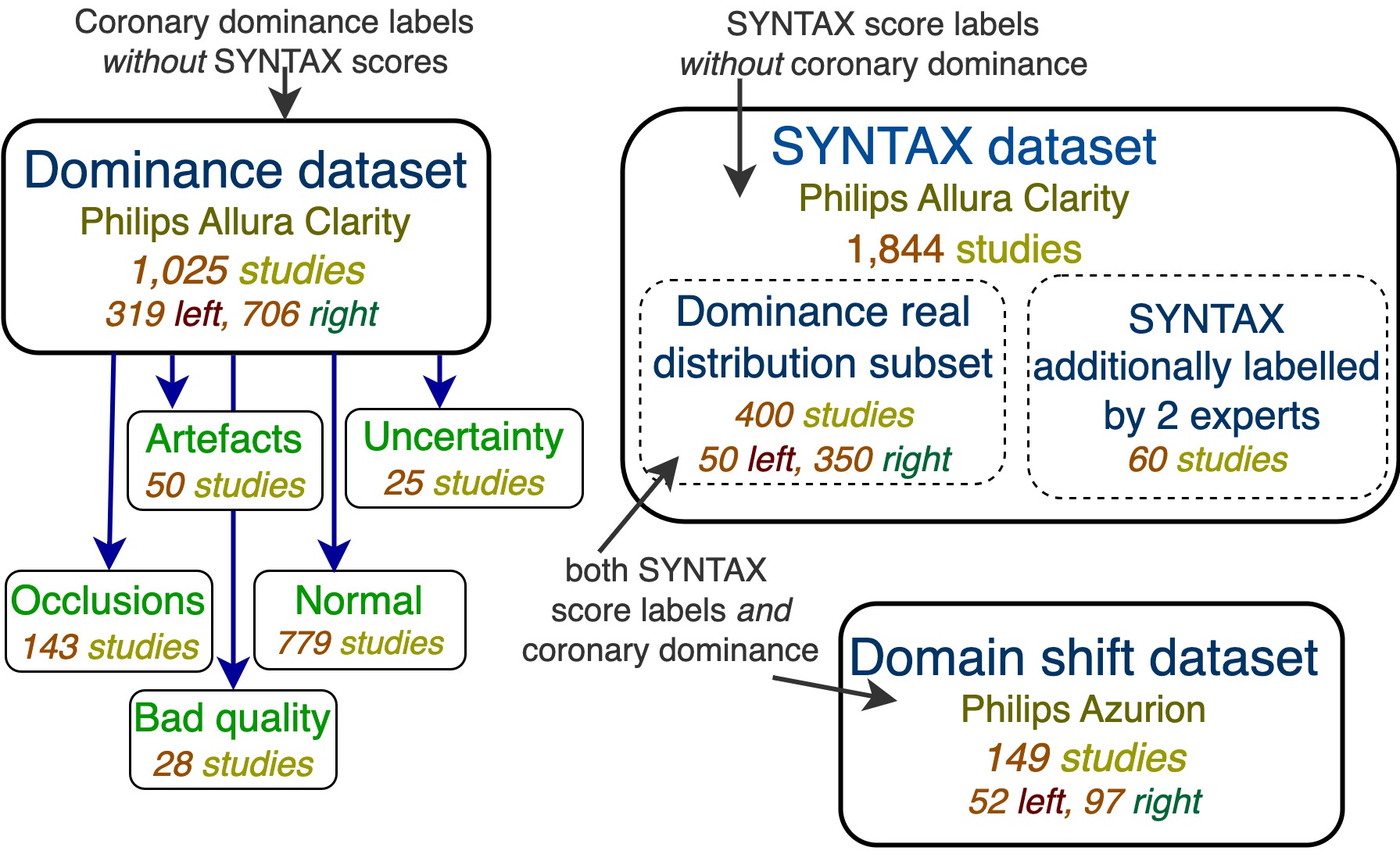}%
\caption{CardioSYNTAX dataset structure.}
\label{fig:dataset_structure}
\end{figure}

The data we present in our CardioSYNTAX dataset includes three parts (Fig. \ref{fig:dataset_structure}): the SYNTAX dataset (Table \ref{table:data_set_size}), the Dominance dataset, and the Domain shift dataset. The total number of angiographic studies is 3,018. We will also call these parts datasets, as they have different labels and can be used separately. However, all these parts consist of invasive coronary angiograms. The SYNTAX dataset is located at \href{https://zenodo.org/records/14005818}{\url{https://zenodo.org/records/14005818}}.

The SYNTAX dataset has 1,844 studies. All of the studies have a SYNTAX score that was estimated by an interventional cardiologist. In addition, 60 of the 1,844 studies in the SYNTAX dataset have scores from two different interventional cardiology specialists, which allows for estimating inter-expert variability. Of the 1,844 studies, 400 were randomly selected and had an additional tag for the coronary dominance.

The Dominance dataset with 1,025 studies (319 of whom have left dominance, and 706 of whom have right dominance) from the Philips Allura Clarity cardiovascular imaging system provides coronary dominance labels, but no SYNTAX score. For the Dominance dataset, each study falls into one of five categories: bad quality, artifact, small diameter, occlusion, or normal. "Normal" is the largest category. We enhanced the main data set with additional left dominant cases and occlusions in order to address the problem of imbalance.

The Domain shift dataset comprises 149 studies (52 left-dominant and 97 right-dominant) from the Philips Asurion cardiovascular imaging system and has both coronary dominance and SYNTAX labels. 

This retrospective study included patients who underwent invasive coronary angiography at the Tyumen Cardiology Research Center
between January 2016 and November 2023\footnote{The study complied with the principles of the Helsinki Declaration, and the protocol was approved by a local ethics committee (Meeting minutes Extract № 184, dated from 16.03.2023).}. The collection procedure for invasive coronary angiography followed the most recent guidelines, utilizing an interventional angiography system, Philips Allura Clarity and Philips Azurion, with an iopromide contrast agent and a frame rate of 15 fps.

An angiographic study of a patient consists of coronary angiography (CA) views (projections), which are medical imaging tools that allow scientists to observe the body in a 3D space and capture it in 2D image planes. The CA projections are not synchronized temporally. This technique uses X-ray technology to produce video images of the internal organs and blood vessels. The examination includes LCA and RCA projections. On average, there are 1-3 RCA projections and 3-5 LCA projections. A single CA projection contains approximately 20-60 frames, with pixels linearly normalized to the range 0-255. Automatic annotation of RCA and LCA images is performed rather than manual classification, as the classification task has an accuracy above 99\%~\cite{eschen2022classification,moalla2023exploiting}.

\begin{table}
\begin{center}
\caption{SYNTAX dataset characteristics}
\label{table:data_set_size}
{\small{
\begin{tabular}{@{}lllc@{}}
\toprule
Number of & LCA & RCA & Total\\
\midrule
Patients  & -N/A- & -N/A- & 1,844\\
\makecell[l]{Angiographic \\ views} & 9,590 & 3,970 & 13,560\\
Frames & 469,557 & 201,591 & 671,148\\
\bottomrule
\end{tabular}
}}
\end{center}
\end{table}

For further information on the size of the SYNTAX dataset, please refer to Table~\ref{table:data_set_size}, which provides information on the number of angiographic views, frames, and split by RCA and LCA.

The average age of patients in the dataset is 63.6 years. Additionally, 59\% of patients in the study were male, and 41\% were female. The patients are from the Tyumen region of Russia. Nationality is not included in the statistics collected at clinics. According to the latest census, in the region 90\% of the population has European origin, while 10\% are Asian or mixed heritage.

The SYNTAX dataset contains an equal number of cases with a zero and non-zero SYNTAX score. This property allows us to predict the SYNTAX score and classify patients with zero and non-zero scores. It is a significant advantage of our dataset, as invasive angiography typically refers to people with a CAD condition~\cite{scanlon1999acc}. The SYNTAX score distribution is shown in Table~\ref{table:syntax_distrib}. The dataset includes the total SYNTAX score for a patient and separate scores for the RCA and LCA vessels. It also consists of a tag if a patient had a coronary artery bypass (187 patients).

\begin{table}
\begin{center}
\caption{SYNTAX score distribution}
\label{table:syntax_distrib}
{\small{
\begin{tabular}{@{}lllc@{}}
\toprule
Score & RCA & LCA & Total\\
\midrule
0 & 88.3\% & 65.0\% & 52.0\% \\
\midrule
1 & 0.12\% & 0.50\% & 0.38\% \\
\midrule
2-5 & 6.12\% & 9.2\% & 13.5\% \\
\midrule
6-10 & 4.70\% & 7.97\% & 9.6\% \\
\midrule
11-20 & 0.75\% & 9.39\% & 12.1\% \\
\midrule
20+ & 0.03\% & 7.92\% & 12.4\% \\
\midrule
Max & 23.0 & 61.0 & 67.0 \\
\bottomrule
\end{tabular}
}}
\end{center}
\end{table}

\subsection{Data labeling} \label{subsection:disagreement}
For the SYNTAX dataset three interventional cardiologists, each with 3 to 15 years of experience, participated in the data labeling process. Each cardiologist labeled only their own portion of the data. For the Dominance and the Domain shift dataset six interventional cardiologists with experience ranging from 1 to 25 years provided ground truth information on coronary dominance. A moderator PhD-level cardiologist with more than 15 years of experience verified their labels. We considered his labels to be a gold standard. He also labeled studies with additional tags, such as occlusions and artifacts. 

We conduct the following experiment to estimate an order of magnitude of experts' disagreement in SYNTAX score labeling. Two interventional cardiologists (one with ten and the other with three years of experience) labeled 60 studies sampled from our dataset. The patients were selected randomly, but only a fifth were selected from the subgroup with zero scores. We compared the scores pairwise and computed $R^{2}$, bias, and deviation for each score pair. The results of the comparison are in Table~\ref{subsection:disagreement}. Bias and deviation are for the Bland-Altman plot. The $R^2$ in experts' disagreement varies between each other and the dataset is between 0.6 and 0.8, which is a high value considering the importance of the score for risk assessment.

\begin{table}
\begin{center}
{\small{
\caption{Experts disagreement for 60 random patients. Bias and deviation are the mean and STD values of the Bland-Altman plot.}
\label{table:expert_disagreemetn}
\begin{tabular}{@{}llllc@{}}
\hline\noalign{\smallskip}
metric & $R^{2}$ & \makecell{Bias \\ mean} & \makecell{Bias \\ median} & \makecell{Deviation \\ STD} \\ 
\noalign{\smallskip}
\hline
\noalign{\smallskip}
\makecell{Expert 1 vs.\\ dataset}  & 0.593 & -1.80 &-1.00 & 9.52\\ 
\hline
\noalign{\smallskip}
\makecell{Expert 2 vs.\\ dataset}  & 0.829 & 0.53 &0.00 &  6.69\\ 
\hline
\noalign{\smallskip}
\makecell{Expert 2 vs.\\ Expert 1}  & 0.674 & 1.27 &1.50 & 9.16\\ 
\hline
\noalign{\smallskip}
Average$_{\pm std}$ & {0.72$_{\pm 0.10}$} & 0.00$_{\pm 1.60}$ &-0.17$_{\pm 1.26}$ & 8.46$_{\pm 1.54}$ \\
\bottomrule
\end{tabular}
}}
\end{center}
\end{table}

\section{Multi-view look method}
\begin{figure}[!t]
\centering
\includegraphics[height=4.5cm]{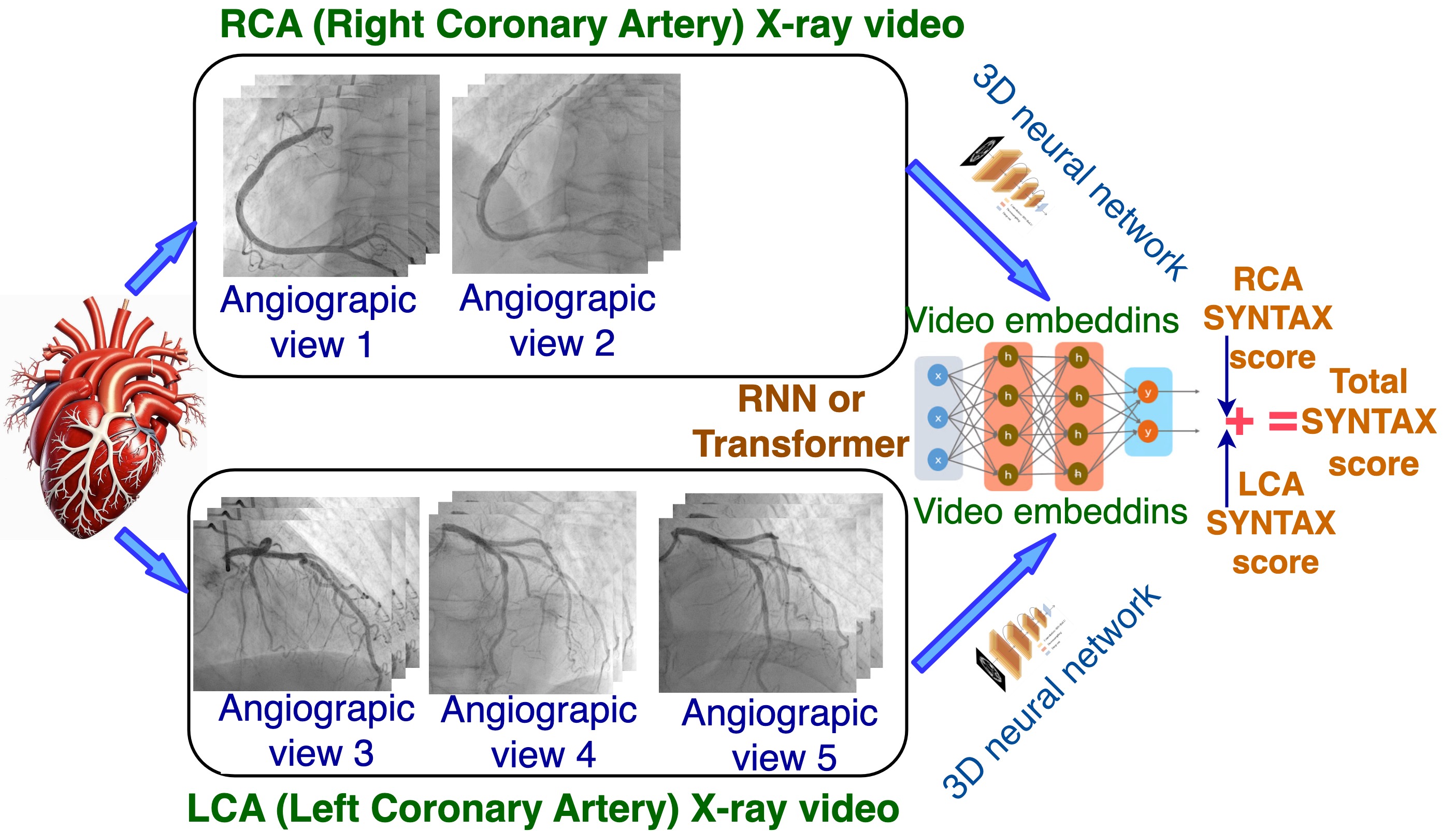}
\caption{Proposed end-to-end SYNTAX score estimation scheme.}
\label{fig:estimation_scheme}
\end{figure}

Our research presents a new method \textendash \space Multi-View Look for predicting SYNTAX scores using a combination of 3D ResNet~\cite{tran2018closer} and recurrent neural networks (RNN) or Transformer models. As shown in Fig.~\ref{fig:estimation_scheme}, our method involves extracting features from each view using a 3D backbone, fusion them with a head neural network, and then utilizing a fully connected layer for classification and regression. Notably, we have employed separate neural networks for predicting the scores of the RCA and LCA views, as their domains differ significantly. The total SYNTAX score is obtained by summing the individual scores of RCA and LCA.

We released the code for SYNTAX score estimation at \href{https://github.com/palevas/syntax-score-prediction}{\url{https://github.com/palevas/syntax-score-prediction}}.

\subsection{Models and training details for SYNTAX score estimation}
Our training process consisted of 3 main steps: backbone training, head training and joint backbone and head fine-tuning. In the first step, we only predicted the SYNTAX score for one angiographic view using the backbone network. In the second step, we freeze the backbone and train a head to predict outcomes from multiple views. At the final step, we fine-tune both the backbone and the head.

We extract features not from a frame, but from a whole video. For this purpose we have utilized an 18-layer 3D ResNet model with approximately 33 million parameters, pre-trained on the KINETICS400~\cite{kay2017kinetics} dataset. Our experiments have shown that using pre-trained weights has significantly improved the prediction quality. To ensure the accuracy of MVL method, we have used a 32-frame video as input, and any missed frames in the angiographic view are filled with repeated frames. We have employed the pytorch lightning framework to train our model on A100 GPU efficiently.

There are various methods for combining embeddings. The early work~\cite{lin2018nextvlad} used NetVlad aggregation for the frame features aggregation. The study~\cite{meng2019frame} used an attention mechanism to combine embeddings from different frames in a video. Liu et al., in their work~\cite{liu2022petr}, used a 3D position encoder and a Transformer decoder to combine information from multi-view images. Dynamic neural radiance fields~\cite{li2022neural} solve the problem of combining 3D multi-view videos to render photorealistic images from various viewpoints and at arbitrary times. However, the position of the angiographic view is loosely defined and can not help for precise 3D reconstruction.

The study \cite{tan20213d} is the most similar to our approach. The difference is that, the authors of \cite{tan20213d} used one 3D image (without a time axis) as input and we had multiple 2D video inputs.
The task of \cite{tan20213d} was to allow for an arbitrary number of input slices, while our approach focused on an arbitrary number of videos from different viewpoints. The authors of \cite{tan20213d} used 3D patches as input, and our network has multiple 2D video inputs. The study \cite{tan20213d} used BERT for information aggregation and we got the best results with LSTMs.

Our research used BERT~\cite{devlin2018bert} and LSTM-RNN to merge characteristics from different angiographic perspectives. BERT head had 1.2 million parameters, while LSTM had only 0.26 million parameters. The problem specific allows us not to use positional encoding for BERT and feed embeddings in arbitrary sequencing. When we didn't apply a head network, we averaged the embedding vectors.

\begin{figure}[!t]
\centering
\includegraphics[height=4.5cm]{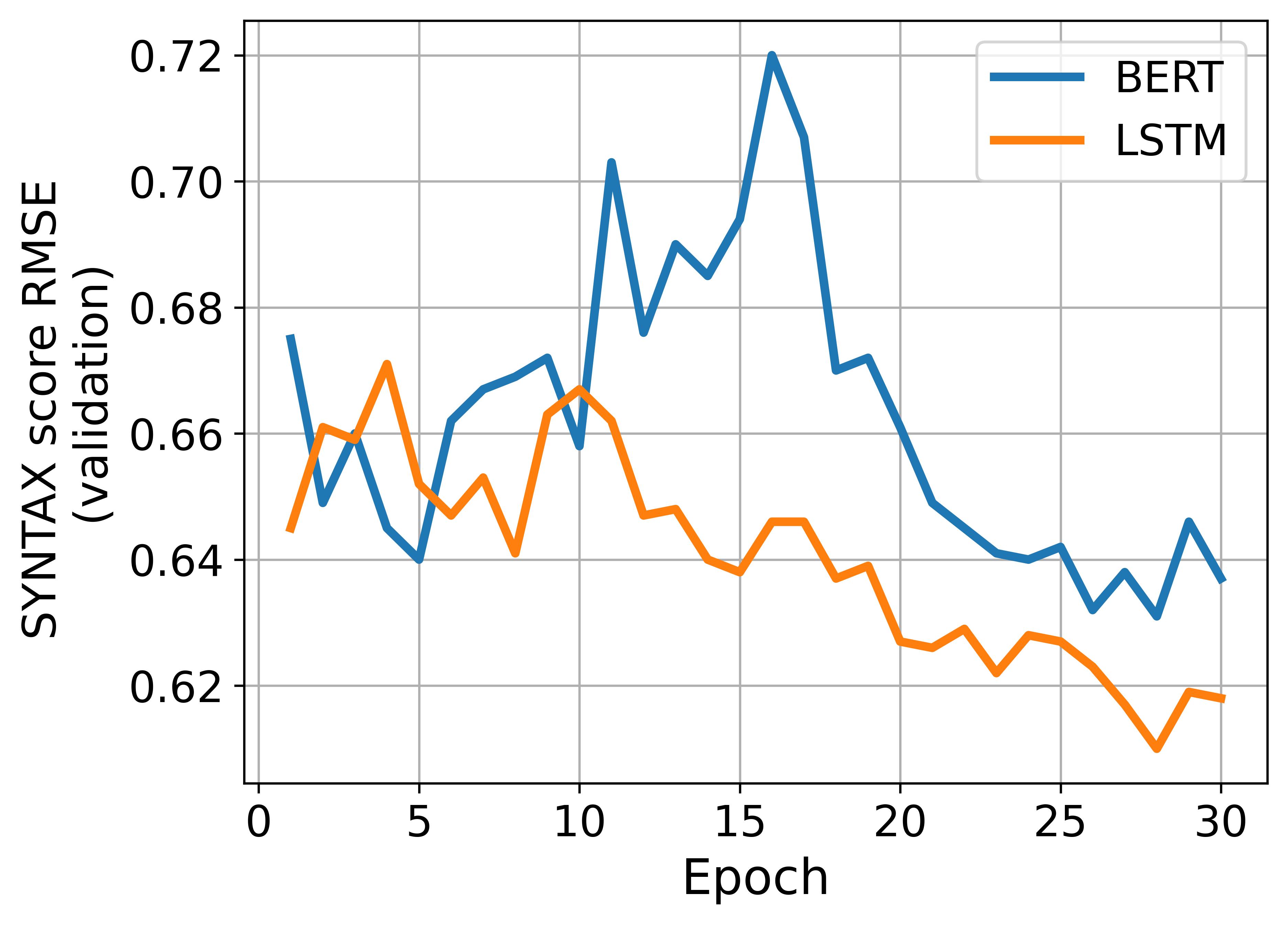}
\caption{Example of RMSE loss (Log of SYNTAX score)} for LSTM and BERT head on validation.
\label{fig:loss_lstm_bert}
\end{figure}

During training, we utilized the OneCycleLR scheduler with a batch size of 16 for backbone training and 8 for head training. The learning rate for training was $10^{-4}$ and $10^{-5}$ for fine-tuning. We used MSE loss in all of our experiments.

\subsection{Ablation study}
The SYNTAX score prediction results using of our framework for different embedding aggregation strategies are presented in Table \ref{table:result_all}. These results demonstrate an advantage of using a head network over simple averaging for end-to-end SYNTAX prediction. An example of training process for LSTM and Bert heads are in Fig. \ref{fig:loss_lstm_bert}.

In recent years, there has been a significant increase in the use of advanced models, such as Video Swin Transformer~\cite{liu2022video} and TubeViT~\cite{piergiovanni2023rethinking}, for various 3D computer vision tasks. However, our study of these models' performance compared to a classical ResNet 3D model as a backbone has yielded surprising results. Surprisingly, despite their state-of-the-art designs, the ResNet 3D model outperformed these advanced models. Upon further analysis, we hypothesize that the small size of our dataset may have played a crucial role in this outcome. It is well-known that transformer models excel in larger datasets but may not be as efficient for smaller-scale datasets~\cite{dosovitskiy2020image}. In our study, we also experimented with an advanced convolution transformer-like model, 3D MedNext~\cite{roy2023mednext}, but it could not surpass the performance of ResNet 3D. The fact could be attributed to the absence of pre-trained weights and the need for further modifications for classification purposes.

Our research used the small Video Swin Transformer, a state-of-the-art model with 28 million parameters pre-trained on the KINETICS400 dataset. This model was readily available through Torchvision. Additionally, we explored using the TubeViT\footnote{https://github.com/daniel-code/TubeViT} model, which boasts 86 million parameters and was pre-trained on the UCF101 dataset~\cite{soomro2012ucf101}. To further investigate the performance of MVL methods, we also experimented with the MedNext\footnote{https://github.com/MIC-DKFZ/MedNeXt} model, which initially operated as an image-to-image model. However, to suit our specific needs (classification vs. segmentation), we removed the decoder component of the MedNext model, resulting in a model with 12 million parameters.

Using the RNN or Transformer head significantly improved the quality of predictions over an average of each ResNet Video output. The results in Table~\ref{table:result_all} show the difference. Besides the LSTM model, we have also tried GRU, but the metrics are left behind LSTM.

\section{Results}
We used the following metrics for SYNTAX score estimation quality \textendash~$R^{2}$ coefficient, bias (mean, median), and deviation (STD, IQR) in terms of the Bland-Altman plot for ground truth and predicted scores. We used standard metrics for the classification quality evaluation \textendash recall, precision, F1, accuracy, and MCC (Matthew correlation~\cite{chicco2020advantages}). The results for the SYNTAX score prediction using our framework are presented in Table~\ref{table:result_all}. These results demonstrate an advantage of using a head network over an averaging strategy and show a clear potential for automatic end-to-end SYNTAX prediction.

\begin{table}
\begin{center}
\caption{SYNTAX score prediction $_{\pm STD}$. 5 fold cross-validation. Bias and deviation are for the Bland-Altman plot.}
\label{table:result_all}
{\small{
\begin{tabular}{@{}lllllc@{}}
\hline\noalign{\smallskip}
method & $R^{2}\uparrow$  & \makecell{Bias \\ mean $\downarrow$}  & \makecell{Bias \\ median$\downarrow$} & \makecell{deviation \\ STD$\downarrow$} \\ 
\noalign{\smallskip}
\hline
\noalign{\smallskip}
\makecell{LSTM \\ (Ours)}  & \textbf{0.512}$_{\pm 0.028}$ &-1.70$_{\pm \textbf{0.32}}$ &\textbf{0.03}$_{\pm 0.05}$ & \textbf{7.51}$_{\pm 0.51}$ \\
\hline
\noalign{\smallskip}
Bert  & 0.448$_{\pm 0.047}$ &-1.65$_{\pm 1.90}$ &  0.17$_{\pm 0.30}$ & 7.79$_{\pm 0.74}$ \\ 
\hline
\noalign{\smallskip}
\makecell{average \\ strategy}  &0.038$_{\pm 0.187}$ & \textbf{-0.99}$_{\pm 0.65}$ &   0.19$_{\pm 0.09}$ &10.74$_{\pm 1.61}$ \\ 
\bottomrule
\end{tabular}
}}
\end{center}
\end{table}

\begin{figure*}
\centering
\subfloat[]{\includegraphics[height=3.1cm]{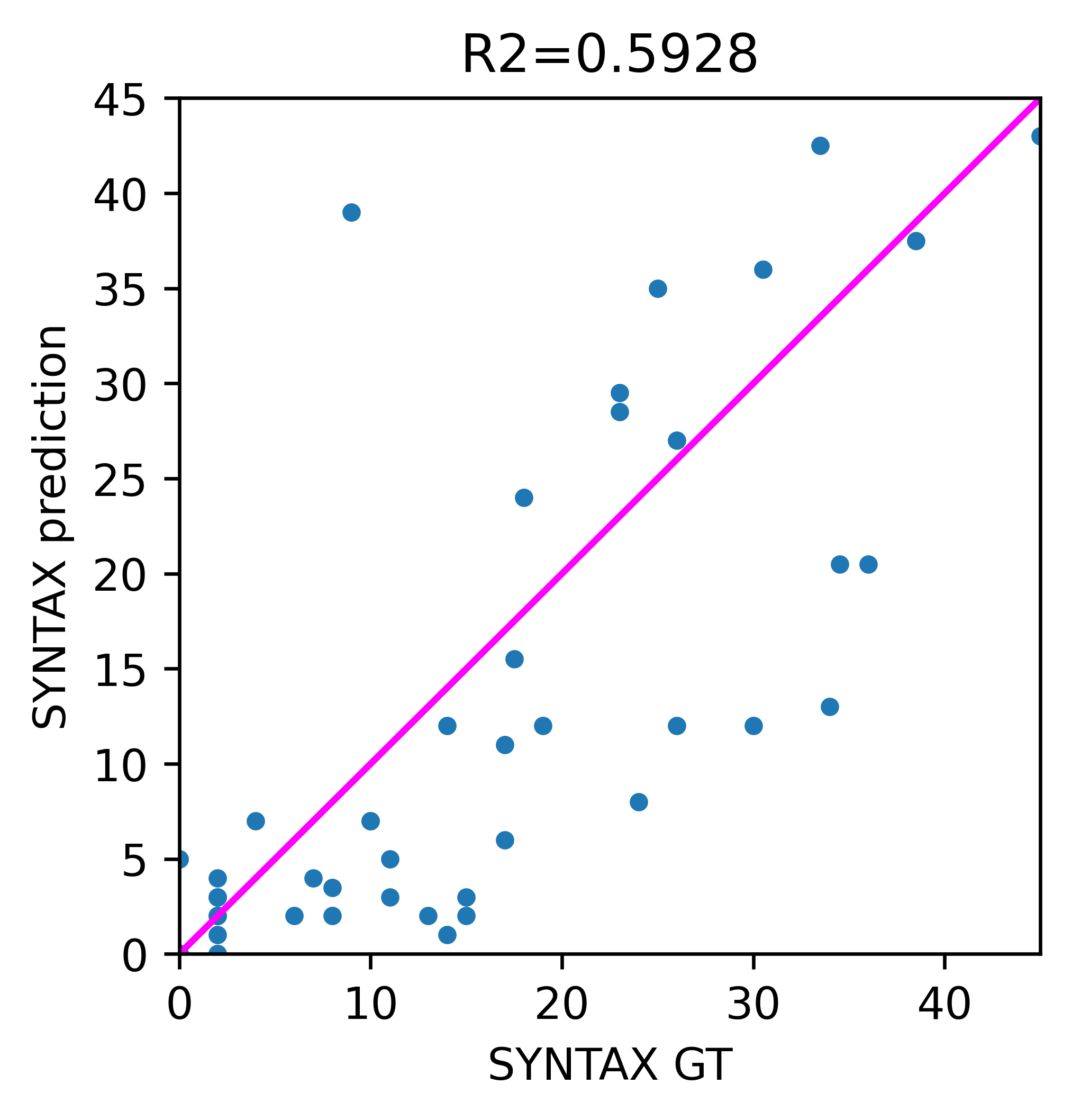}\label{fig_stenosis}}
\subfloat[]{\includegraphics[height=3.1cm]{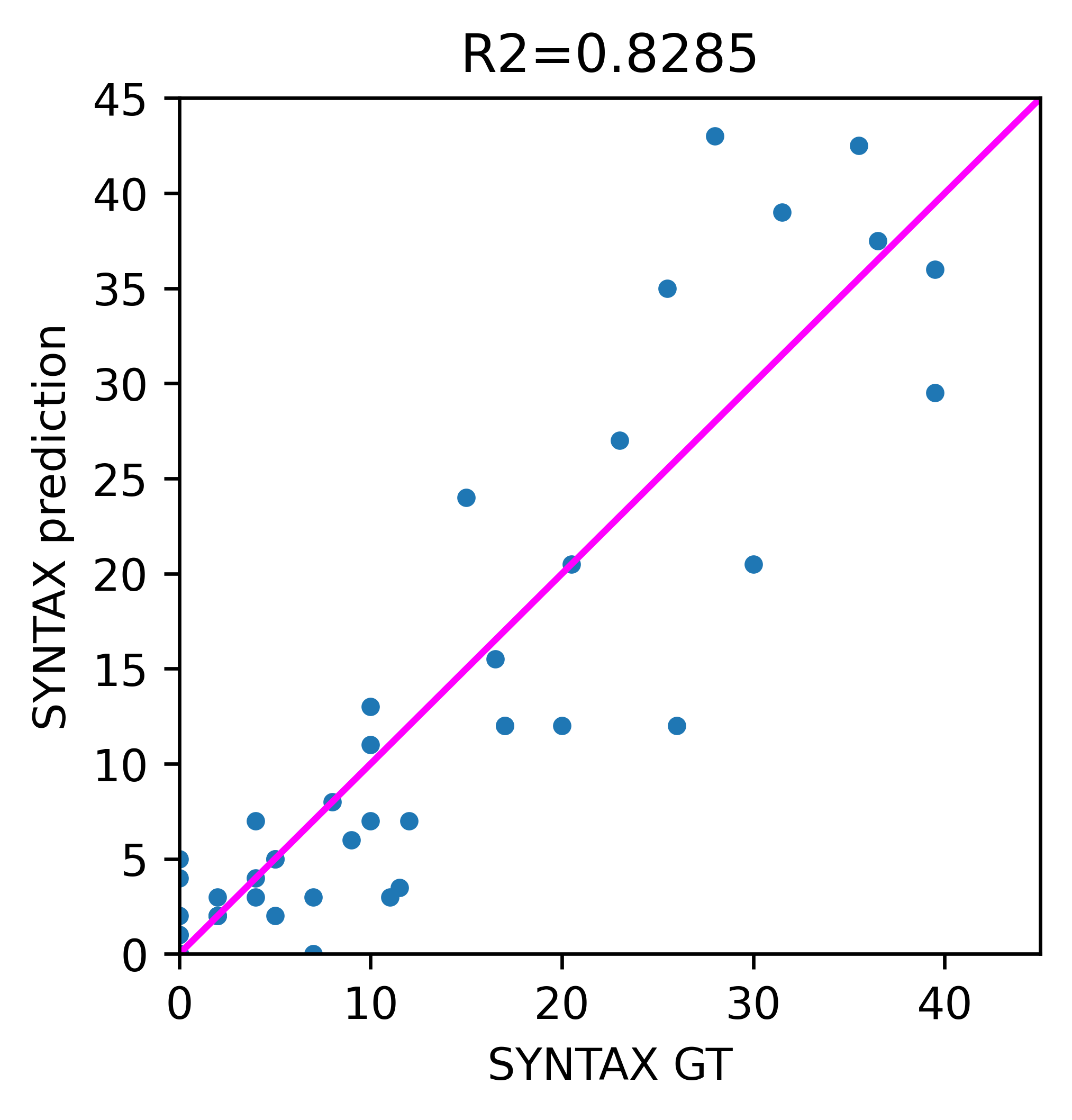} \label{fig_bifurcation}}
\subfloat[]{\includegraphics[height=3.1cm]{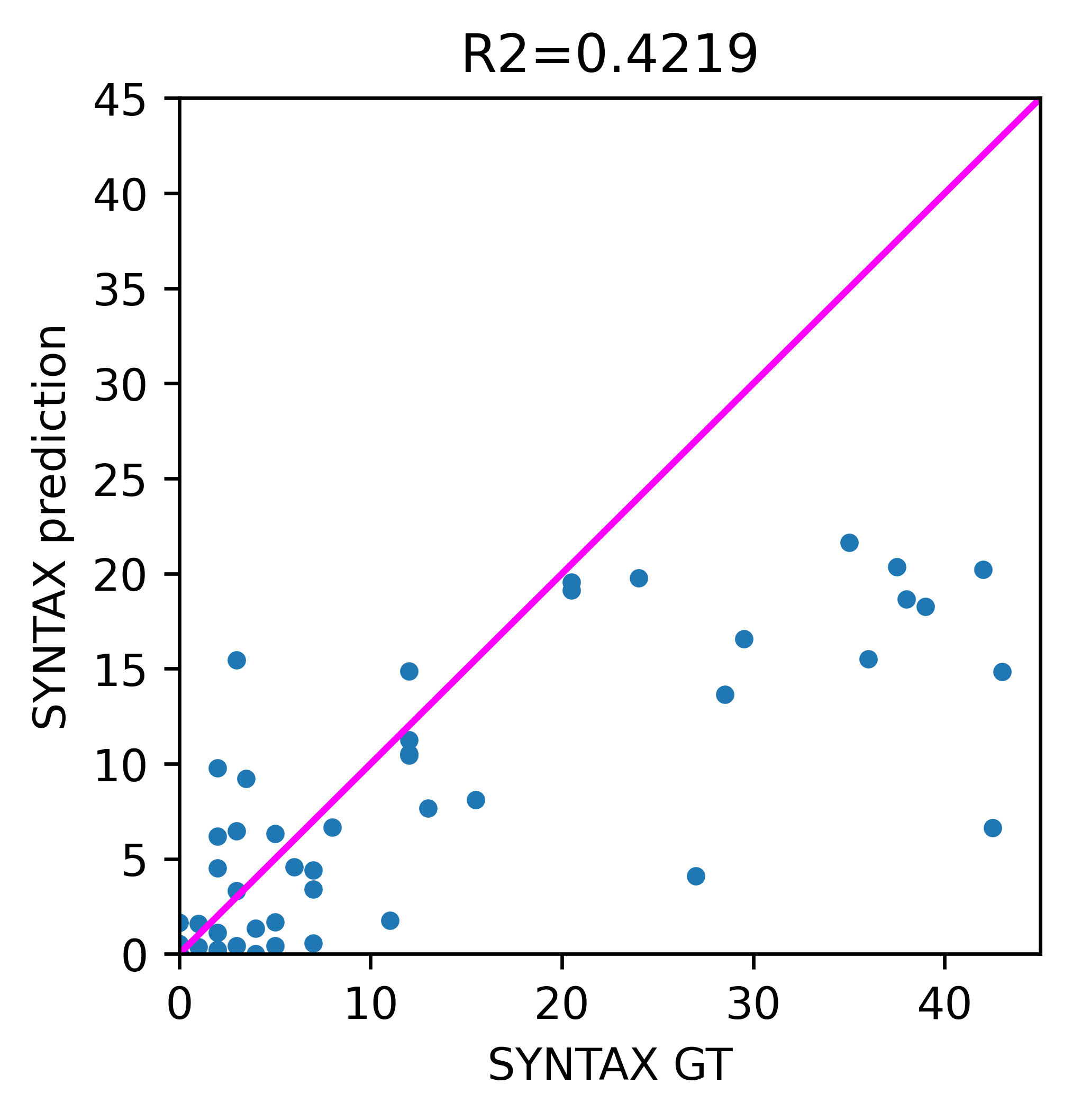}
\label{fig_calcific}}
\subfloat[]{\includegraphics[height=3.1cm]{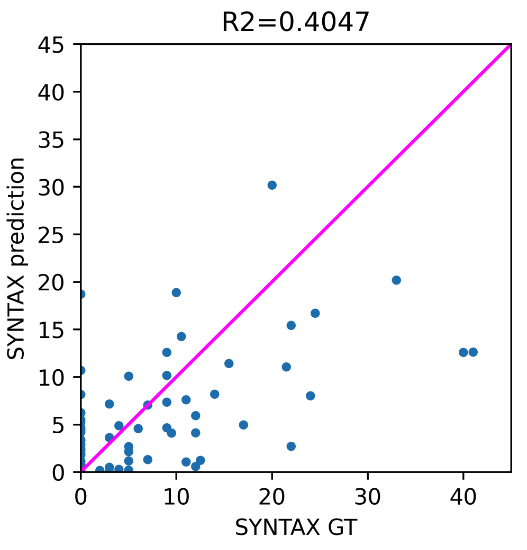}
\label{fig_domain_shift_r2}}
\caption{Correlations plots for SYNTAX score prediction in 60 patients selected at random (a) Expert 1 vs. dataset. (b) Expert 2 vs. dataset. (c) Our prediction (LSTM head) vs. SYNTAX dataset. (d) Our prediction (LSTM head) vs. Domain shift dataset}
\label{fig:R2}
\end{figure*}

We have also separated the prediction results for sick persons with non-zero SYNTAX scores in Table~\ref{table:result_non_zero_score}. The motivation to split the results into two categories came from the fact that the number of healthy patients may differ significantly from country to country, as national standards have different recommendations and contraindications for invasive angiography~\cite{scanlon1999acc}. For this reason, the share of patients with the zero score may also differ among datasets in different clinics. The results in Table~\ref{table:result_non_zero_score} show a worse performance than Table~\ref{table:result_all} and demonstrate how sensitive the results are to changes in patient score distribution. This issue should be addressed in future research.

\begin{table}
\begin{center}
\caption{SYNTAX score prediction $_{\pm STD}$ tested on the patients with ground truth SYNTAX above zero. 5-fold cross-validation. Bias and deviation are for the Bland-Altman plot.}
\label{table:result_non_zero_score}
{\small{
\begin{tabular}{llll}
\toprule
method & LSTM  & BERT &\makecell{Average \\ strategy} \\
\midrule
$R^{2}\uparrow$   & \textbf{0.226}$_{\pm 0.045}$ & 0.154$_{\pm 0.091}$ &-0.61$_{\pm 0.26}$ \\
\midrule
Bias mean $\downarrow$  & \textbf{-4.78}$_{\pm 0.46}$ & -4.97$_{\pm 2.76}$ & -8.56$_{\pm 1.11}$ \\
\midrule
Bias median$\downarrow$  &\textbf{-3.46}$_{\pm0.41}$ & -3.80$_{\pm 1.86}$ &   -3.48$_{\pm 0.70}$ \\
\midrule
Deviation STD$\downarrow$  & \textbf{9.58}$_{\pm 0.78}$ & 9.62$_{\pm 1.12}$ &  15.06$_{\pm 2.43}$ \\
\bottomrule
\end{tabular}
}}
\end{center}
\end{table}

We used 5-fold cross-validation to obtain metrics in Tables~\ref{table:result_all},~\ref{table:result_non_zero_score},~\ref{table:zero_classification}. The cross-validation split used for Table~\ref{table:result_non_zero_score} is the same as that used for Table \ref{table:result_all}, with the only difference being that we used only studies with non-zero scores during testing.
\begin{table}
\begin{center}
\caption{SYNTAX $>$ 0 classification. 5 fold cross-validation $\pm$STD. }
\label{table:zero_classification}
\centering
{\small{
\begin{tabular}{@{}lllc@{}}
\hline
\thead{Metric} & \thead{LSTM} &\thead{BERT} &\thead{average \\ strategy} \\
\hline \makecell{Recall $=$ 0 $\uparrow$ \\ (Sensitivity)} & 74.4$_{\pm\textbf{6.2}}$\% &\textbf{75.41}$_{\pm14.0}$\% &$62.3_{\pm7.3}$\% \\
\hline \makecell{Recall $>$ 0 $\uparrow$ \\ (Specifity)} & 80.4$_{\pm5.9}$\% &79.3$_{\pm8.3}$\% & \textbf{87.9}$_{\pm5.2}$\%\\
\hline \makecell{Precision $=$ 0 $\uparrow$} & 80.5$_{\pm5.4}$\% &80.5$_{\pm5.2}$\% &\textbf{84.8}$_{\pm5.8}$\%\\
\hline \makecell{Precision $>$ 0 $\uparrow$} & 74.6$_{\pm5.1}$\% &\textbf{76.4}$_{\pm7.9}$\% &$68.5_{\pm4.4}$\%\\
\hline \makecell{Recall$_{macro}$ $\uparrow$} &  \textbf{77.4}$_{\pm4.9}$\% &\textbf{77.4}$_{\pm5.2}$\% &$75.1_{4.9\pm}$\%\\
\hline \makecell{F1$_{macro}$ $\uparrow$} & \textbf{77.3}$_{\pm 4.9}$\% &77.0$_{\pm5.8}$\% &74.3$_{\pm5.1}$\%\\
\hline \makecell{MCC $\uparrow$} & 55.0$_{\pm9.7}$\% &\textbf{55.8}$_{\pm9.3}$\% &51.7$_{9.6\pm}$\%\\
\hline \makecell{Accuracy $\uparrow$} & \textbf{77.3}$_{\pm4.9}$\% &\textbf{77.3}$_{\pm5.4}$\% &74.6$_{\pm4.9}$\%\\
\hline
\end{tabular}
}}
\end{center}
\end{table}
\setlength{\tabcolsep}{1.4pt}

The classification results are in Table~\ref{table:zero_classification}. MVL method allows us to predict 80\% of the patients with zero SYNTAX, and the precision is about 75\%. If we improve the precision in the future, interventional cardiologists can pay more attention to complex cases and fewer patients with a zero score. Our new dataset, which we present in this paper, contains an equal number of cases with zero and nonzero scores, making it very suitable for regression and classification.

\begin{table}
\begin{center}
\caption{Comparison of our prediction error with the disagreement between doctors. The SYNTAX score prediction was tested on 60 random patients. Bias and deviation are for the Bland-Altman plot.}
\label{table:comparison_with_disagreement}
{\small{
\begin{tabular}{@{}lllllc@{}}
\toprule
method & $R^{2}\uparrow$  & \makecell{Bias \\ mean $\downarrow$}  & \makecell{Bias \\ median$\downarrow$} & \makecell{deviation \\ STD$\downarrow$} \\ 
\noalign{\smallskip}
\midrule
\noalign{\smallskip}
LSTM (our)  & 0.42 &6.13 &1.52 & 10.87 \\
\midrule
Disagreement$_{\pm STD}$  & {\textbf{0.72}$_{\pm 0.10}$} & 0.00$_{\pm \textbf{1.60}}$ &0.17$_{\pm 1.26}$ & \textbf{8.46}$_{\pm 1.54}$ \\ 
\bottomrule
\end{tabular}
}}
\end{center}
\end{table}

Before predicting the SYNTAX score for patients, we first predicted it for LCA and RCA. The $R^{2}$ coefficients for LCA and RCA SYNTAX scores (LSTM head) are 0.46 and 0.25, ROC AUC 85\% and 82\%, and accuracy 78\% and 80\%. This indicates the apparent difficulties with RCA SYNTAX prediction. RCA affects the SYNTAX score less than LCA, but this issue should be addressed in the future.

To assess the effectiveness of our framework prediction, we compared the level of disagreement between experts (Subsection~\ref{subsection:disagreement}). To ensure the robustness of our results, we trained our framework a second time and utilized a test subset of 60 patients from the aforementioned disagreement group. The results of this comparison can be found in Table~\ref{table:comparison_with_disagreement} and are further illustrated in Fig~\ref{fig:R2}.

We tested our 5 models for SYNTAX score estimation on 104 studies of the Domain shift dataset from another cardiovascular imaging system (Philips Azurion) we used for model training. The metrics are $R^2$=0.376$\pm$0.072 (Fig.\,\ref{fig_domain_shift_r2}), bias=-1.35$\pm$0.31, STD=6.57$\pm$0.41, specificity=66.3$\pm$8.1\%, sensitivity=80.5$\pm$1.5\%. The achieved regression and classification quality demonstrate that our model is robust to the domain shift problem.

For coronary dominance classification we achieved $97.3_{\pm0.4}$\% accuracy and $93.9_{\pm0.9}$\% macro F1 on 400 studies from the Dominance real distribution subset of the SYNTAX dataset. For the Domain shift dataset the accuracy was $89.4_{\pm1.6}$\% and macro F1 $88.1_{\pm1.9}$\%.

\subsection{Discussion}
The proposed framework for fully automatic SYNTAX score estimation from coronary angiography has shown promising results, achieving an $R^{2}$ coefficient of 0.51 in score prediction and an accuracy of 77\% in classifying patients with zero and non-zero scores. While this indicates a clear potential for our MVL method, it is essential to note that it still falls behind the estimation of experts, as demonstrated in Table~\ref{table:comparison_with_disagreement}. This is due to the higher correlation between the labels assigned by the experts and the dataset compared to our predictions. Additionally, MVL method has shown a tendency to underestimate scores, as seen in Table~\ref{table:result_all}, which can be attributed to the use of MSE loss in the training process and the presence of many studies with zero scores. Furthermore, the performance of the patients with non-zero scores in Table~\ref{table:result_non_zero_score} may also be affected by this issue, as the model was trained on a dataset where half of the studies have zero scores but tested on studies with non-zero scores.

The head of the neural network is a crucial component in determining the model's performance. In the context of utilizing a 3D backbone, the choice of the head greatly impacts the results obtained. While the LSTM head has shown superior performance in regression tasks, BERT has demonstrated similar performance in classification tasks.

The SYNTAX Score assessment influences the decision-making process regarding patient treatment strategies, specifically the choice between percutaneous coronary intervention procedure and coronary artery bypass grafting surgery. The decision is based on the range into which the Syntax score falls. The most significant ranges are as follows: 0-22 points, 23-32 points, and more than 33 points.  For 60 studies, the average disagreement between experts was $STD=8.5$. This is a high value, and with such a STD, the same angiography study scores can fall into non-overlapping ranges. For example, on Fig.~\ref{fig:R2}c, many scores above 33 were attributed by one expert as being below 22.

\section{Conclusions}
Our research team has proposed a new multi-view 3D regression and classification problem for the AI community, focused on estimating the SYNTAX score from invasive coronary angiography. To further support our proposal, we have curated a comprehensive  CardioSYNTAX dataset of coronary angiograms for 3,018 patients, 1,844 of whom have their corresponding SYNTAX scores and 1,025 of whom have coronary dominance labels. Additionally, 149 studies were collected from another cardiovascular imaging system with SYNTAX and coronary dominance labels. To establish a benchmark for prediction quality, we have also calculated the disagreement between experts in SYNTAX evaluation, with a determination coefficient $R^{2}$ ranging from 0.59 to 0.83. Our proposed method has demonstrated results, with an $R^{2}$ of 0.51 between our predictions and the scores from the dataset.

To enhance the accuracy of our predictions, we have suggested using RNN or Transformer models by concatenating the embeddings from each angiographic view. This approach has significantly improved prediction quality, outperforming the average prediction strategy.

Our findings show a clear prospect for end-to-end SYNTAX score prediction, though the achieved results still need to be more for the algorithm to be used for CAD risk assessment. A presented novel CardioSYNTAX dataset of angiographic studies contributes to developing more accurate and efficient methods for SYNTAX score estimation. The following issues need to be further investigated and addressed:
\begin{itemize}
\item Mitigating the effect of the SYNTAX distribution in the test set on the results, e.g., testing the algorithm on data with non-zero scores and vice versa.
\item Identify complex cases for AI models and discuss how to interpret and improve these issues with cardiologists.
\item Additional research on expert disagreement in data labeling, through increased involvement of new experts in labeling and the number of studies labeled by different experts.
\end{itemize}

\clearpage
%
%
\bibliographystyle{splncs04}
\bibliography{egbib}
\end{document}